\documentclass[letterpaper]{article} 
\usepackage{aaai25}  
\usepackage{times}  
\usepackage{helvet}  
\usepackage{courier}  
\usepackage[hyphens]{url}  
\usepackage{graphicx} 
\usepackage{natbib}  
\usepackage{caption} 
\usepackage{makecell} 
\usepackage{graphicx}
\usepackage{booktabs}
\usepackage[table,xcdraw]{xcolor}
\usepackage{colortbl}
\usepackage{xcolor}
\usepackage{booktabs}
\usepackage{algorithm}
\usepackage{algpseudocode}
\usepackage{multirow}
\usepackage{pifont}
\usepackage{cellspace}

\usepackage{newfloat}
\usepackage{listings}
\DeclareCaptionStyle{ruled}{labelfont=normalfont,labelsep=colon,strut=off} 
\floatstyle{ruled}
\newfloat{listing}{tb}{lst}{}
\floatname{listing}{Listing}
\title{HeGTa: Leveraging Heterogeneous Graph-enhanced Large Language Models for Few-shot Complex Table Understanding}
\author{
    Rihui Jin\textsuperscript{\rm 1,2},
     Yu Li\textsuperscript{\rm 1,2}, 
     Guilin Qi\textsuperscript{\rm 1,7}\thanks{Corresponding author.}, 
     Nan Hu\textsuperscript{\rm 1,2}, 
     Yuan-Fang Li\textsuperscript{\rm 3}, 
     Jiaoyan Chen\textsuperscript{\rm 4}, 
     Jianan Wang\textsuperscript{\rm 5},\\
     Yongrui Chen\textsuperscript{\rm 1,2}, 
     Dehai Min\textsuperscript{\rm 1,2}, 
     Sheng Bi\textsuperscript{\rm 6}
}
\affiliations{
    \textsuperscript{\rm 1}School of Computer Science and Engineering, Southeast University, China\\
    \textsuperscript{\rm 2}Key Laboratory of New Generation Artificial Intelligence Technology and its Interdisciplinary Applications \\(Southeast University), Ministry of Education, China \\
    \textsuperscript{\rm 3}Monash University 
    \textsuperscript{\rm 4}The University of Manchester
    \textsuperscript{\rm 5}Alibaba Group\\
    \textsuperscript{\rm 6}Law and Innovation Lab, Law School, Southeast University, China\\
    \textsuperscript{\rm 7}State Key Laboratory for Novel Software Technology, Nanjing University, China\\
    
    \{ari\_king, yuli\_11, gqi, nanhu, yrchen, zhishanq, bisheng\}@seu.edu.cn, 
    yuanfang.li@monash.edu, 
    jiaoyan.chen@manchester.ac.uk, jn\_wang2@outlook.com
}

\usepackage{bibentry}

\begin{document}

\maketitle

\begin{abstract}
Table Understanding (TU) has achieved promising advancements, but it faces the challenges of the scarcity of manually labeled tables and the presence of complex table structures.
To address these challenges, we propose HeGTa, a heterogeneous graph (HG)-enhanced large language model (LLM) designed for few-shot TU tasks.
This framework aligns structural table semantics with the LLM's parametric knowledge through soft prompts and instruction tuning. 
It also addresses complex tables with a multi-task pre-training scheme by incorporating novel multi-granularity self-supervised HG pre-text tasks. 
We empirically demonstrate the effectiveness of HeGTa, showing that it outperforms the SOTA for few-shot complex TU on several benchmarks.
\end{abstract}

\section{Introduction}
\begin{figure}[!t]
\centering 
  \includegraphics[width= 0.42\textwidth]{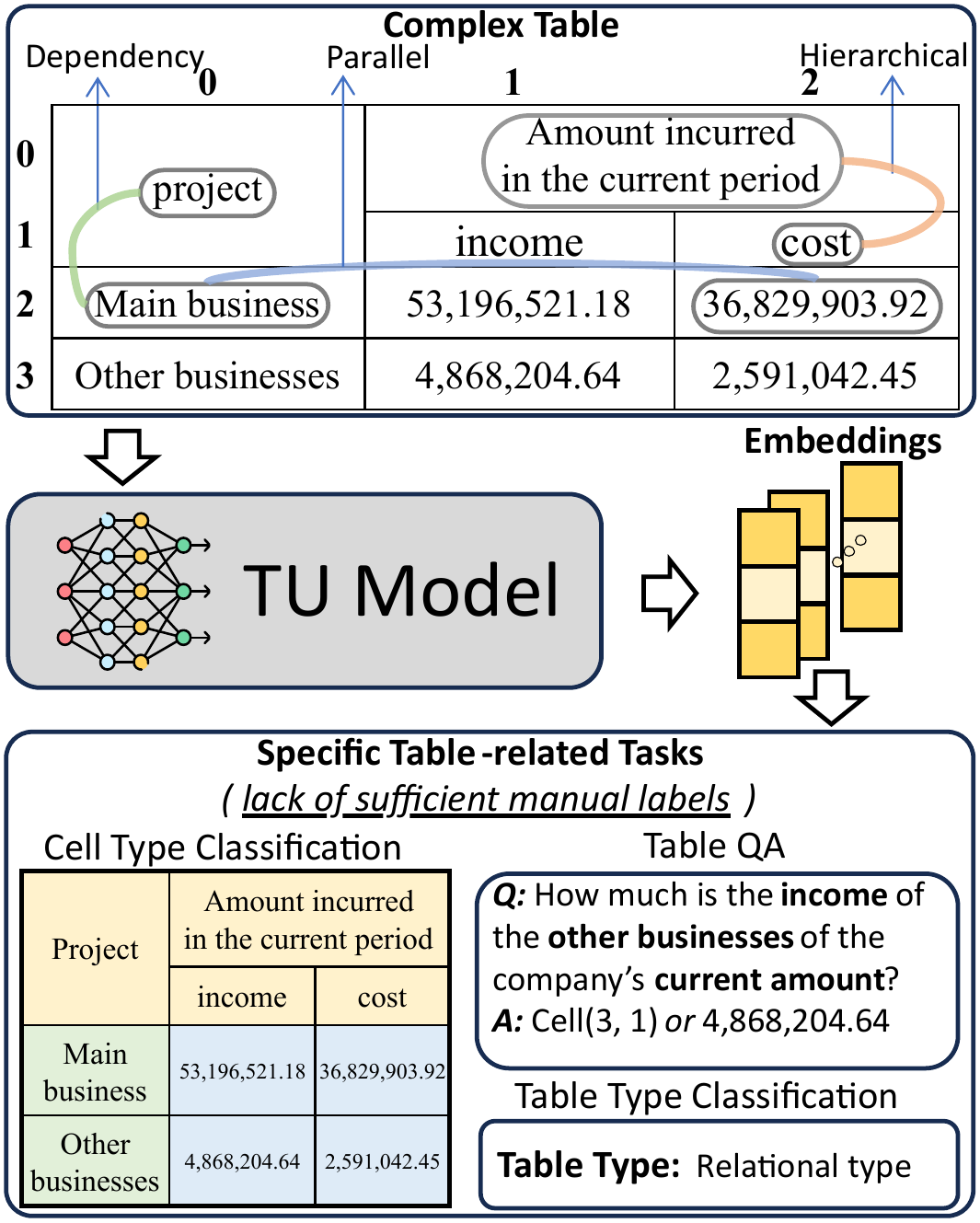}
  \caption{Few-shot complex Table Understanding. Complex tables contain intricate cell-to-cell relationships, including dependency, hierarchical, and parallel ones.}
  \label{fig: intro}
\end{figure}

Table Understanding (TU) seeks to learn informative embeddings of tables containing inherently tabular semantics, often formatted in ways not easily interpretable by machines, as shown in Fig.~\ref{fig: intro}. This endeavor enhances machine performance across a range of table-related tasks, such as Table QA~\cite{chain-of-table}, Cell Type Classification~\cite{getpt} and Table Type Classification~\cite{lu2024large}.

Yet, in real-world scenarios, TU faces the challenges of \textbf{a lack of sufficient human annotations} and the presence of \textbf{complex table structures}, which diminishes the effectiveness and applicability of existing frameworks. As for the first challenge, the data-hungry nature of existing frameworks results in diminished performance in few-shot TU scenarios, where only several samples are annotated. Although some studies~\cite{dongsurvey, tapas} have adopted pre-training with encoder-only architectures~\cite{bert} to alleviate the annotation burden, these solutions still require considerable amounts of labeled data for task-specific fine-tuning.
Regarding the second challenge, while current frameworks attempt to capture structural information in tables through the use of position embeddings~\cite{tuta} or by modeling tables as graphs~\cite{getpt, GTR}, these methods are effective for simple tables, but wane with complex tables. This shortfall arises because the cell-to-cell relationships in complex tables are more intricate than those in simple tables, as shown in Fig.~\ref{fig: intro}.

Fortunately, recent advancements have introduced techniques that can be integrated into the TU framework to tackle the challenges: 
1)  Large Language Models (LLMs) have shown remarkable effectiveness in managing few-shot tasks involving data in other modalities, such as vision~\cite{llava} and time series~\cite{timellm}. 
This is achieved through instruction tuning and soft prompts, which seek to align the semantic spaces of the other modality encoder with that of the LLM, a method that can also be applied to tabular data.
2) Self-supervised heterogeneous graph (HG) pre-training~\cite{sshg1, sshg2} enables models to navigate multifaceted relationships within data, such as complex intercell relationships in Fig.~\ref{fig: intro}, using substantial volumes of unlabeled data.

We propose HeGTa, an HG-enhanced LLM framework for the few-shot complex TU. 
To enable HeGTa to model intricate relationships within complex tables, HeGTa begins with modeling the tabular data with the HG and then processing it through a tabular graph encoder to generate vectors imbued with structural information.
Additionally, to ensure that the LLM achieves comparable performance in table tasks to its performance in natural language (NL) tasks, especially in few-shot scenarios, we align the representation spaces of the tabular graph encoder and the LLM via instruction tuning. 
Specifically, we integrate table vectors containing structural information as soft prompts within the LLM inputs and innovatively design three multi-granularity self-supervised tasks tailored for both tables and LLMs for pre-training.
After pre-training specific parameters of HeGTa via these self-supervised tasks, the HeGTa can capitalize on the LLM's exceptional generalization capability to adapt to downstream tasks with minimal data samples. 
To validate the performance of HeGTa in few-shot complex TU, we conduct extensive comparative experiments with existing powerful baselines on nine publicly available datasets for three specific table-related tasks in few-shot scenarios. The experimental results show that HeGTa exceeds the current SOTA for the few-shot complex TU across multiple benchmark datasets.
In summary, our main contributions are as follows.
\begin{itemize}
\setlength{\topsep}{0pt}
\setlength{\partopsep}{0pt}
\setlength{\itemsep}{0pt}
\setlength{\parsep}{0pt}
\setlength{\parskip}{0pt}

\item To the best of our knowledge, we are the first to propose a framework to align table semantics with NLs' to empower the LLM to perform on tables with the same few-shot proficiency as it does with NLs.
\item To improve the framework's ability to complex TU, we propose a refined way to convert tables into HGs and design three novel multi-granularity self-supervised HG pre-training tasks tailored for tabular data and LLMs.
\item We conduct extensive experiments on nine publicly available datasets, and the experimental results show that HeGTa exceeds the current SOTA for few-shot complex TU across multiple benchmark datasets.

\end{itemize}

\section{Related Work}

\subsection{Graph-based Table Understanding}

Many studies~\cite{tabularnet,tabprompt} have converted tables into graphs and utilized graph encoders to capture tables' inherent topological information. 
These frameworks include employing homogeneous graphs or utilizing a basic node-linking strategy that connects cells exclusively to their adjacent counterparts. 
Consequently, such frameworks underperform when dealing with complex table structures.

\subsection{LLM-based Textual Table Understanding}

Following the success of LLMs in NL tasks, some efforts~\cite{tablellama, llmfew, tableformatsurvey} have extended their application to table-related tasks. 
Although these frameworks leverage the exceptional generalization capacity of LLMs to achieve SOTA performance in some few-shot tasks, they resort to simply converting the table into a row-by-row NL format as the input. This process leads to a loss of the tables' intrinsic topological information.

\subsection{LLM-based Symbolic Table Understanding}
Inspired by Program-of-Thoughts\cite{chen2022program}, several studies~\cite{api, Binding,chain-of-table,ReAcTable,dater} have adapted semantic parsing techniques, traditionally applied to relational database tables, to general Table QA tasks. This involves transforming the table into a format interpretable by query languages, such as SQL, and subsequently utilizing LLMs to generate a symbolic program to retrieve the answer. This methodology represents the current SOTA in Table QA. However, its architectural limitations restrict its applicability to other table-related tasks.

\subsection{Bert-like Encoder-only Table Understanding}
Since the rise in popularity of pre-training models like BERT~\cite{bert}, there has been considerable effort devoted to designing specialized encoding methods for tabular data and unique pre-text objectives for pre-training~\cite{tabert, fortap, getpt}. 
Despite their utilization of self-supervised training, these methods still require a substantial amount of labeled data during fine-tuning for downstream tasks. Additionally, while they incorporate positional embedding into serialized tabular data, they do not effectively capture topological information.

\section{Task Definition}
Given a table $T=\{c_{i,j} | 0\leq i < N, 0 \leq j < M\}$ where $N$ is the number of rows, $M$ is the number of columns, and $c_{i,j}$ is the cell located in the $i^{th}$ row and $j^{th}$ column.
Merged cells, characterized by a row span or column span greater than 1, are prevalent in complex tables. For instance, the cell labeled "project" in the top-left corner of the table shown in Fig.~\ref{fig:t2g} has a row span of 2. We assign the coordinates of such merged cells based on the position of the top-left cell before merging. Hence, the coordinate of the "project" cell is designated as (0,0).

\noindent\textbf{Specific sub-TU tasks for Evaluation} 

Cell Type Classification (CTC) involves identifying the type $y_c$ of each cell $c_{i,j}$ within a table $T$, where $y_c$ can belong to a basic taxonomy $Y_c=\{header\ cell, data\ cell\}$ or a more complex one, varying across datasets.

Table Type Classification (TTC) is a table-level categorization task that requires models to classify the table $T$ according to a specific taxonomy $Y_t$.

Table QA (TQA) demands that the model produce an answer $y_a$ in response to a natural language question $q$, with table $T$ serving as the reference for deriving the answer $y_a$.

\section{METHODOLOGY}

\begin{figure*}[h]
    \centering 
      \includegraphics[width=0.82\textwidth]{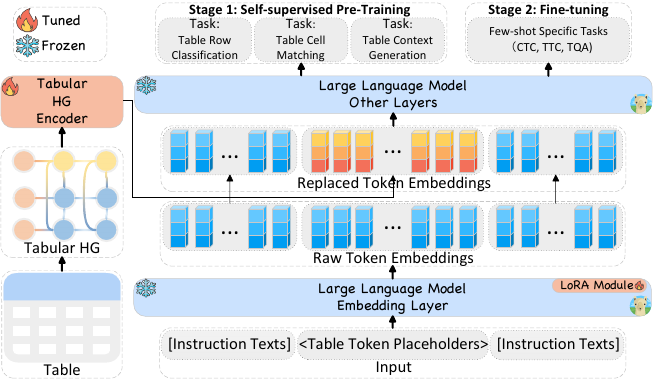}
      \caption{An overview of HeGTa framework. HeGTa processes \textless table, instruction\textgreater ~as an input. First, the table is converted into an HG and processed by a Tabular HG encoder to generate a vector for each tabular node, while the LLM transforms instruction texts into initial token embeddings. Subsequently, the HG encoder's outputs serve as soft prompts for the LLM, enabling the replacement of placeholder embeddings with actual tabular node vectors. The modified embedding sequence is then processed by the remaining LLM layers. Throughout Stage 1 and Stage 2, only the weights of red components are tuned.}
      \label{fig: overview}
\end{figure*}

This section explains three phases of HeGTa: tabular HG construction and two tuning stages, as shown in Fig.~\ref{fig: overview}.

\subsection{Tabular Heterogeneous Graph Construction}
\label{subsection: Heterogeneous Tabular Graphs}

\begin{figure*}[h]
\centering 
    \includegraphics[width=0.681\textwidth]{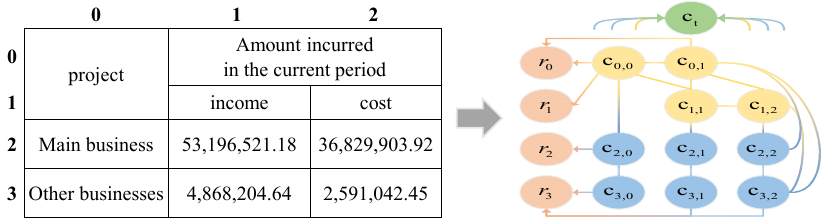}
    \caption{Table-to-heterogeneous graph conversion. Node types are color-coded: Table (green), Row (red), Data Cell (blue), and Header Cell (yellow). Edge types are similarly color-coded, with bidirectional edges shown as undirected lines. Some edges are omitted for clarity.}
    \label{fig:t2g}
\end{figure*}

Given that HGs are more proficient at capturing diverse relationships compared to homogeneous graphs, we employ HGs with heuristic node-linking rules to model the structure of complex tables effectively. The subsequent subsections detail the process of converting tabular data into HGs, including the creation of nodes and the heuristic rules for establishing edges between nodes. The process of creating a tabular HG is shown in Fig.~\ref{fig:t2g}.

\subsubsection{Four Tabular Node Types} 
TABLE, ROW, Header CELL, and Data CELL nodes are denoted as green, red, yellow, and blue nodes in Fig.~\ref{fig:t2g}. 1) The TABLE node represents the content described by the table, facilitating the table-level tasks. 2) The ROW node signifies the information contained within a row, aiding in the prediction of the row's type during self-supervised training. 3) The Header CELL node denotes cells located in the header row, identifying column schemas or categories. 4) The Data CELL node represents cells in data rows, meaning the actual data entries of the table.

\subsubsection{Creating Tabular Nodes}
First, CELL nodes are created for each cell in the table, with each node denoted as $c_{i,j}$, where $i$ and $j$ represent the cell's coordinates in the original table. If a CELL node is located in the table's header section or is part of a merged cell spanning the entire table width, it is viewed as a Header Cell; otherwise, it is identified as a Data Cell. The algorithm for determining the header section is elaborated in Appendix. Subsequently, ROW nodes are created to match the number of rows in the table, along with a single TABLE node.

\subsubsection{Initializing Tabular Node Embeddings}
An initialization vector is required for each node in the HG. For Header and Data CELL nodes, we employ the output of S-BERT~\cite{sentencebert} applied to the text within each cell to obtain their initialization vectors. 
In the case of ROW nodes $r_i$, we initialize a vector by concatenating the text from the cells in the $i^{th}$ row and inputting this concatenated text into the S-BERT. 
TABLE nodes serve to represent the table's content itself. In a human's view, a table's content can be inferred by examining the cells in the headers. Consequently, we opt to concatenate the text from the cells in the headers and to obtain the embedding.

\subsubsection{Adding Edges}
To enhance the machine's understanding of the table's semantics, we develop the following heuristic rules to link nodes:
1) A TABLE node, representing the whole table, should be linked to all CELL nodes to encapsulate the global semantics.
2) A ROW node derives information from cells within the same row, so the ROW node $r_i$ should be linked to each CELL node $c_{i*}$.
3) A strong correlation exists between the semantics of Data CELLs and their corresponding Header CELLs within the same column, so they should be linked. 
4) Data CELLs located in the same column exhibit a stronger relational bond compared to those in different columns. Consequently, adjacent Data CELLs within a column are interconnected.
5) All Header CELLs should be interconnected because the interpretation of relationships between Data CELLs across columns necessitates the semantic understanding of their respective Header CELLs. For instance, as demonstrated in Fig.~\ref{fig:t2g}, determining the relationship between cell $c_{2,0}$ and cell $c_{2,1}$ requires an examination of their Header CELLs. This examination reveals that ``53,196,521.18'' represents the income from the ``main business'' project.

Edge types are categorized based on the nodes they link, as follows: Table-Header, Table-Data, Header-Row, Data-Row, Header-Data, Data-Data, and Header-Header edges.

\subsection{Stage 1: Self-supervised Instruction Tuning}

Illustrated in Fig.~\ref{fig: overview}, HeGTa utilizes the tabular output of the HG encoder as soft prompts~\cite{softprompt}, which are part of the LLM input. The weights of both modules are tuned through self-supervised instruction tuning to align the vector representation spaces of the two modules. This subsection provides details of the pre-training process, which involves three different granularity self-supervised tasks.

\subsubsection{Tabular HG Encoder}

Following the conversion of the table into a tabular HG, HeGTa introduces a heterogeneous GNN (HGNN)~\cite{HGT} as the encoder for tabular HG data. The encoder takes a tabular HG as input and generates vector representations for the tabular nodes as output. The HGNN employs a message-passing mechanism to collect semantic and topological information from neighboring cell nodes within each layer, considering various edge types. 


\subsubsection{Multi-granularity Self-supervised Instruction Tasks}
Drawing inspiration from the methodologies in LLaVA~\cite{llava}, Time-LLM~\cite{timellm}, and GraphGPT~\cite{graphgpt}, we develop a method to effectively align the vector spaces of two distinct modalities of data: tables and NLs. 
We introduce the soft prompt technique as a bridge between the two encoders and focus on fine-tuning the tabular graph encoder and the LoRA~\cite{lora} module within the LLM.
This process, designed to be lightweight, enables the LLM to grasp the topological information of tables through semantic instruction tuning.

\begin{figure*}[h]
\centering 
    \includegraphics[width= 0.81\textwidth]{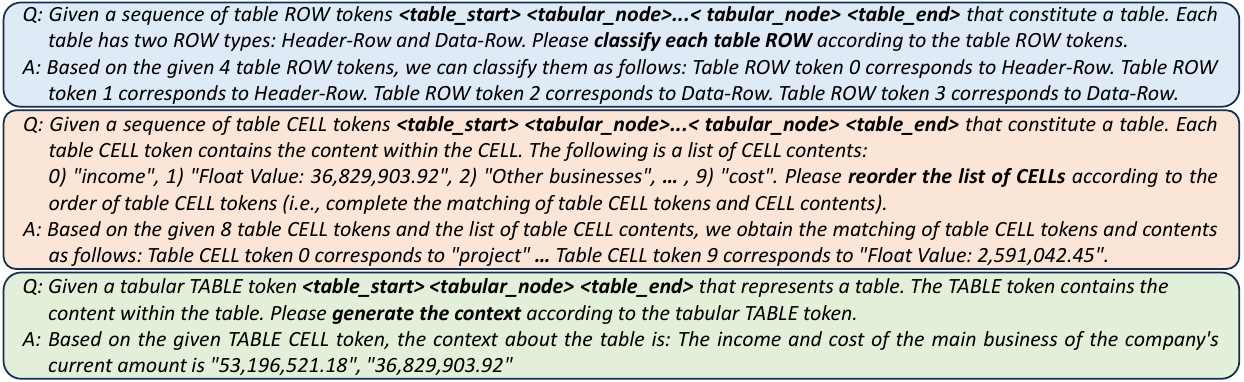}
    \caption{Examples of three self-supervised instruction tuning datasets, each tailored for distinct tasks: Table Row Classification, Table Cell Matching, and Table Context Generation.}
    \label{fig:datasample}
\end{figure*}

Instruction tuning~\cite{instruct1, instruct2}, a technique that merges fine-tuning with prompt learning, significantly enhances the generalization capabilities of LLMs. Therefore, before the model-tuning process, we pre-format the self-supervised training data into an instruction-based QA format. Examples of training data for instruction tuning are illustrated in Fig.~\ref{fig:datasample}. 

To align the LLM with the HG encoder, the HeGTa incorporates additional tokens into the vocabulary: \textless tabular\_node\textgreater, \textless table\_start\textgreater~and \textless table\_end\textgreater. The token \textless tabular\_node\textgreater~serves as a placeholder for table tasks within the instruction text, allowing for the substitution with actual table node vectors post-processing by the LLM's Embedding layer. 
The quantity of these placeholders matches the number of table nodes relevant to the current task. 
The tokens \textless table\_start\textgreater~and \textless table\_end\textgreater~signify the beginning and ending delimiters of the table placeholders. 

The forward propagation in HeGTa begins with the input of a \textless Table, Instruction\textgreater~pair, where the instruction text passes through the LLM's embedding layer, and the table is transformed into a Tabular HG before being processed by the HGNN. 
The LLM embedding layer assigns embeddings to each token, creating a sequence, while the HGNN's output provides aggregated vector representations for each table node. Subsequently, HeGTa replaces the embeddings corresponding to the table token placeholders within the sequence with the actual node vectors. This adjusted embedding sequence is then fed into the LLM's remaining layers.

To enhance the HeGTa's understanding of tables, we introduce three tasks tailored for the graph-enhanced LLM, each varying in granularity: Table Row Classification, Table Cell Matching, and Table Context Generation.
These tasks are designed to integrate table representations into the LLM's semantic space. The complexity of these tasks increases progressively, aiming to incrementally improve the model's ability to comprehend semantic information. Detailed descriptions of each task are provided below.

\noindent\textbf{Table Row Classification (TRC).}
The objective of this task is to train the model to accurately identify the category of each table row. Utilizing the vector representations of Row nodes provided by the HG encoder, the model discerns whether a node corresponds to a Header Row or a Data Row. The categorization of Row nodes is automated by an algorithm elaborated in Appendix, eliminating the need for manual intervention and thus qualifying as a self-supervised training process. In the dataset we utilize, the labeling accuracy of the algorithm reaches 97.6\%. This task facilitates the model's initial grasp of coarse-grained information about the structure of table rows.

\noindent\textbf{Table Cell Matching (TCM).}
The task involves supplying the model with vector representations of each cell node alongside a list of shuffled cell contents. The model's objective is to correctly pair each cell node vector with its corresponding original text. For instance, referring to the table depicted in Fig.~\ref{fig:t2g}, the model needs to align the node vector of $c_{1,1}$ with the string "income" within the list correctly. This training task enables the model to discern the semantic information encapsulated within the cell contents based on the graph node vectors. Essentially, it aligns the semantic space of the HG encoder and the LLM.

\noindent\textbf{Table Context Generation (TCG).}
This task aims to enable the model to generate context surrounding table data, utilizing the vector representations from the Table node. This task facilitates the model's learning of the table's global semantic information, proving beneficial for tasks requiring a comprehensive understanding of the table as an entity.

Examples of datasets in instruction format for the three tasks are provided in Fig.~\ref{fig:datasample}.

\subsection{Stage 2: Task-specific Instruction Fine-tuning} 
Following the completion of the self-supervised pre-training tasks in Stage 1, the HeGTa has successfully aligned the representation spaces of the HG encoder and the LLM. This alignment, combined with the intrinsic capabilities of the LLM, empowers HeGTa to effectively comprehend the topological nuances of complex tables. 
Consequently, when applied to a specific downstream task (i.e., CTC, TTC, TQA), HeGTa requires only a small number of training samples to grasp the expected answer format for the downstream task and reorganize its pre-existing knowledge into the suitable output format. Example prompts of downstream tasks can be found in Appendix.
When fine-tuning in Stage 2, the model adopts the same tuning strategy as in Stage 1, with all parameters frozen except for those of the HG Encoder.
To enhance the reasoning capabilities of HeGTa, we annotated the reasoning process with Chain of Thought (CoT)~\cite{wei2022chain} for $k$-shot TQA samples. Inspired by Auto-CoT~\cite{zhang2022automatic}, we first clustered the table samples in the test set, and then clustered the corresponding questions within each table cluster. Finally, we selected the central TQA samples from each question cluster for annotation. This approach ensures the diversity of contextual examples with a limited number of annotations.

\section{Experiments}

\begin{table*}[hbt]
  \centering

    \begin{tabular}{|cccccccccccccc|}
    \toprule
        \multicolumn{1}{c}{\multirow{2}{*}{Datasets}}                 & \multicolumn{3}{c}{Annotation Type}                                                                                                                &  \multicolumn{4}{c}{Table Info} \\
        \cmidrule(lr){2-4} \cmidrule(lr){5-8}
        \multicolumn{1}{c}{}                                          & CTC                                                 & TTC                                             & \multicolumn{1}{c}{TQA}               &  \# Test Tables                               &   \% Complex Tables                       &   \# QA pairs                & \multicolumn{1}{c}{Main Domains} \\ \midrule[0.5pt]
        \multicolumn{1}{l}{IM-TQA}	                                  & \multicolumn{1}{c}{\ding{51}}                       &	\multicolumn{1}{c}{\ding{51}}                   &	\multicolumn{1}{c}{\ding{51}}              &	\multicolumn{1}{c}{153}                     &	\multicolumn{1}{c}{47.71}               &	\multicolumn{1}{c}{627}           &	\multicolumn{1}{c}{Manufacturing}	                    \\
        \multicolumn{1}{l}{WCC}	                                      & \multicolumn{1}{c}{}                                &	\multicolumn{1}{c}{\ding{51}}                   &	\multicolumn{1}{c}{}                       &	\multicolumn{1}{c}{371}                     &	\multicolumn{1}{c}{--}                  &	\multicolumn{1}{c}{--}            &	\multicolumn{1}{c}{General}	                          \\
        \multicolumn{1}{l}{HiTab}	                                    & \multicolumn{1}{c}{\ding{51}}                       &	\multicolumn{1}{c}{}                            &	\multicolumn{1}{c}{\ding{51}}              &	\multicolumn{1}{c}{3597}                    &	\multicolumn{1}{c}{92.88}               &	\multicolumn{1}{c}{1584}          &	\multicolumn{1}{c}{Crime, Health}	                    \\
        \multicolumn{1}{l}{WTQ (Flatten)}	                            & \multicolumn{1}{c}{}                                &	\multicolumn{1}{c}{}                            &	\multicolumn{1}{c}{\ding{51}}              &	\multicolumn{1}{c}{2108}                    &	\multicolumn{1}{c}{0.00}                &	\multicolumn{1}{c}{4344}          &	\multicolumn{1}{c}{General}	                          \\
        \multicolumn{1}{l}{WTQ (Raw)}	                                & \multicolumn{1}{c}{}                                &	\multicolumn{1}{c}{}                            &	\multicolumn{1}{c}{\ding{51}}              &	\multicolumn{1}{c}{2108}                    &	\multicolumn{1}{c}{14.80}               &	\multicolumn{1}{c}{4344}          &	\multicolumn{1}{c}{General}	                          \\	
        \multicolumn{1}{l}{TabFact}	                                  & \multicolumn{1}{c}{}                                &	\multicolumn{1}{c}{}                            &	\multicolumn{1}{c}{\ding{51}}              &	\multicolumn{1}{c}{1695}                    &	\multicolumn{1}{c}{0.00}               &	\multicolumn{1}{c}{12779}         &	\multicolumn{1}{c}{General}	                          \\	\bottomrule
      \end{tabular}
  
  \caption{Dataset Statistics. ``\ding{51}'' indicates the type of annotation in the dataset that has this task in it. ``\# Test Tables'', ``\% Complex Tables'' and ``\# QA pairs'' columns show the number of tables in the test set, the percentage of complex tables and the numbers of QA pairs, respectively.}
  \label{table: Dataset_Statistics}
\end{table*}

\newcommand{\smallsuperscript}[1]{\textsuperscript{\smaller[2]#1}}
\begin{table*}[hbt]
\centering
  \setlength{\tabcolsep}{1.mm}
  \begin{tabular}{ccccccccccc}
  \toprule
  \multicolumn{1}{c}{\multirow{2}{*}{Model Type}}	                                  &  \multicolumn{1}{c}{\multirow{2}{*}{Models}}	            & \multicolumn{2}{c}{CTC }			                                                                                   & \multicolumn{2}{c}{TTC }		                                                                      & \multicolumn{5}{c}{TQA }				              \\ 	\cmidrule(lr){3-4} \cmidrule(lr){5-6} \cmidrule(lr){7-11} 
  \multicolumn{1}{l}{}	                                                            &  \multicolumn{1}{l}{}	                                    & \multicolumn{1}{c}{HiTab}	                                         & IM--TQA	                                   & \multicolumn{1}{c}{WCC}	                          & {IM--TQA}	                                  & \multicolumn{1}{c}{WTQ (Flat)}	                & {WTQ (Raw)}	             & {TabFact}	            & {HiTab}	                  & {IM--TQA}	            \\ 	\midrule[0.5pt]
  \multicolumn{1}{c}{\multirow{1}{*}{Bert-like}}	                                  &  \multicolumn{1}{l}{ForTap}	                              & \multicolumn{1}{c}{56.33            }                              & 47.34                                       & \multicolumn{1}{c}{46.14}	                        & 53.40                                       & \multicolumn{1}{c}{27.33 }	                    & 26.52                    & 28.67                  & 24.88                     & 29.08	                \\
  \multicolumn{1}{c}{\multirow{1}{*}{Encoder-based}}                                &  \multicolumn{1}{l}{GetPt}	                              & \multicolumn{1}{c}{58.45            }                              & 49.67                                       & \multicolumn{1}{c}{48.62}	                        & 56.45                                       & \multicolumn{1}{c}{23.69 }	                    & 23.34                    & 25.03                  & 23.34                     & 22.07	                \\  \midrule[0.5pt]
  \multicolumn{1}{c}{\multirow{2}{*}{Graph-based}}                                  &  \multicolumn{1}{l}{TabularNet}	                          & \multicolumn{1}{c}{54.14            }                              & 45.62                                       & \multicolumn{1}{c}{46.01}	                        & 56.41                                       & \multicolumn{1}{c}{19.18 }	                    & 18.58                    & 20.81                  & 22.34                     & 20.05	                \\
                                                                                    &  \multicolumn{1}{l}{TabPrompt}	                          & \multicolumn{1}{c}{\underline{63.56}}                              & 46.77                                       & \multicolumn{1}{c}{50.31}	                        & 54.34                                       & \multicolumn{1}{c}{17.71 }	                    & 14.77                    & 19.56                  & 20.20                     & 14.14	                \\  \midrule[0.5pt]
  \multicolumn{1}{c}{\multirow{3}{*}{\makecell{LLM-based \\ (Symbolic) }}}	        &  \multicolumn{1}{l}{Binder}	                              & \multicolumn{1}{c}{--               }                              & --                                          & \multicolumn{1}{c}{--}	                            & --                                          & \multicolumn{1}{c}{43.16 }	                    & 40.05                    & 59.04                  & 44.11                     & 41.25	                \\
                                                                                    &  \multicolumn{1}{l}{Aug-Codex}	                          & \multicolumn{1}{c}{--               }                              & --                                          & \multicolumn{1}{c}{--}	                            & --                                          & \multicolumn{1}{c}{33.32 }	                    & 32.28                    & 57.07                  & 45.87                     & 42.16	                \\  
                                                                                    &  \multicolumn{1}{l}{COTable}	                            & \multicolumn{1}{c}{--               }                              & --                                          & \multicolumn{1}{c}{--}	                            & --                                          & \multicolumn{1}{c}{\underline{44.38}}	          & \underline{42.97}        & \underline{59.76}      & \underline{49.68}         & 43.06	                \\  \midrule[0.5pt]
  \multicolumn{1}{c}{\multirow{2}{*}{\makecell{LLM-based \\ (Textual)}}}            &  \multicolumn{1}{l}{TableCoT}	                            & \multicolumn{1}{c}{53.12            }                              & 49.76                                       & \multicolumn{1}{c}{45.57}	                        & 53.80                                       & \multicolumn{1}{c}{31.42 }	                    & 32.12                    & 52.01                  & 47.55                     & 45.21	                \\
                                                                                    &  \multicolumn{1}{l}{TableLlama}	                          & \multicolumn{1}{c}{60.15            }                              & \underline{53.40}                           & \multicolumn{1}{c}{\underline{51.56}}              & \underline{57.55}                           & \multicolumn{1}{c}{38.88 }	                    & 39.50                    & --                     & --                        & \underline{48.82}     \\  \midrule[0.5pt]
  \multicolumn{2}{c}{\multirow{1}{*}{HeGTa}}	                          	                                                                      & \multicolumn{1}{c}{\textbf{66.39}   }                              & \textbf{59.58}                              & \multicolumn{1}{c}{\textbf{55.32}}                 & \textbf{58.46}	                            & \multicolumn{1}{c}{\textbf{45.12}}	            & \textbf{45.96}    	     & \textbf{60.61}	        & \textbf{51.56}	          & \textbf{50.73}        \\	\bottomrule
  \end{tabular}

  \caption{Overall evaluation results on three TU tasks with best \textbf{bolded} and runner-up \underline{underlined}.  
  `--' indicates that the current framework cannot handle the current task due to some limitations. \protect\footnotemark  } 
  \label{table: Overall performance of the HeGTa.}
\end{table*}

\subsection{Datasets}
\label{sec: datasets}
To evaluate HeGTa, we selected a variety of datasets that are widely studied and easy to parse, including \textit{WCC}~\cite{wcc}, \textit{IM-TQA}~\cite{imtqa}, \textit{HiTab}~\cite{hitab}, \textit{WTQ (Raw)} along with its header-flatten version \textit{WTQ (Flatten)}~\cite{wtq} and \textit{TabFact}~\cite{tabfact}, each pertinent to CTC, TTC, or TQA, respectively.

Statistics for these datasets are presented in Table~\ref{table: Dataset_Statistics}, illustrating the types of annotations, the primary domains covered, and the proportion of complex tables. Given the focus on few-shot TU scenarios, we only list the size of test sets. Comprehensive details of the dataset annotations and pre-processing are available in Appendix.

\subsection{Baselines}
We compare HeGTa with several strong baselines to verify its effectiveness. These baselines can be categorized into four groups according to their frameworks.
ForTap~\cite{fortap} and GetPt~\cite{getpt} emulate BERT, devising specific pretext tasks tailored for tabular data to pre-train the Transformer encoder~\cite{bert}. 
TabularNet~\cite{tabularnet} and TabPrompt~\cite{tabprompt} incorporate a graph encoder. 
Binder~\cite{Binding}, Aug-Codex~\cite{api} and COTable~\cite{chain-of-table} leverage LLMs to generate symbolic programs, which are then executed to obtain the final answer.
Compared to symbolic methods, TableLlama~\cite{tablellama} and TableCoT~\cite{llmfew} leverage the textual understanding capabilities of LLMs to directly address tabular tasks.

\subsection{Implementation Details}

\noindent\textbf{General Setup Across All Tasks.} 
We employ Vicuna-7B-v1.5~\cite{vicuna} as the base LLM and a 2-layer Heterogeneous Graph Transformer (HGT)~\cite{rgat} as our tabular HG encoder whose hidden dimension is set as 1024. The initial vectors of nodes in HG are obtained by S-BERT~\cite{sentencebert}, whose dimension is 768. 
We integrate a LoRA~\cite{lora} module to the embedding layer of the LLM. At any tuning stage, HeGTa exclusively tunes the parameters of the HGT and the LoRA module. This lightweight configuration allows our model to be trained on a single 4090 GPU. 
We assemble the tables from training and validation sets of the datasets mentioned above along with a selection of \textit{TURL}~\cite{turl} corpus as a comprehensive dataset consisting of 200k tables for use in the pre-training stage.

\begin{table*}[!ht]
  \centering
  \begin{tabular}{ccccccccccc}
  \toprule
  \multicolumn{1}{c}{\multirow{2}{*}{Module}}	              & \multicolumn{2}{c}{CTC }			                                                                                   & \multicolumn{2}{c}{TTC }		                                                                                      & \multicolumn{5}{c}{TQA }				                                                          \\ 	\cmidrule(lr){2-3} \cmidrule(lr){4-5} \cmidrule(lr){6-10} 
  \multicolumn{1}{l}{}	                                    & \multicolumn{1}{c}{HiTab}	                                         & IM--TQA	                                   & \multicolumn{1}{c}{WCC}	                                          & {IM--TQA}	                                  & \multicolumn{1}{c}{WTQ (Flat)}	                      & {WTQ (Raw)}	                                              & {TabFact}	                      & {HiTab}	                  & {IM--TQA}	            \\ 	\midrule[0.5pt]
  \multicolumn{1}{l}{$\hspace{0.15cm} w/o$\ TRC}            & \multicolumn{1}{c}{-2.65}                              & -1.53                           & \multicolumn{1}{c}{+1.02}	                            & -1.40                           & \multicolumn{1}{c}{-1.10}	                & \multicolumn{1}{c}{-1.34}	                    & -0.89               & -0.78         & +1.19     \\
  \multicolumn{1}{l}{$\hspace{0.15cm} w/o$\ TCM}            & \multicolumn{1}{c}{-4.50}                              & -4.88                           & \multicolumn{1}{c}{-3.22}	                            & -2.44                           & \multicolumn{1}{c}{-2.31}	                & \multicolumn{1}{c}{-2.71}	                    & -2.10               & -1.67         & -2.45     \\  
  \multicolumn{1}{l}{$\hspace{0.15cm} w/o$\ TCG}            & \multicolumn{1}{c}{+0.11}                              & +1.19                           & \multicolumn{1}{c}{-1.61}	                            & -1.73                           & \multicolumn{1}{c}{-2.21}	                & \multicolumn{1}{c}{-2.54}	                    & -2.57               & -0.30         & -2.58     \\  
  \multicolumn{1}{l}{$\hspace{0.15cm} w/o$\ HG}             & \multicolumn{1}{c}{-3.28}                              & -1.61                           & \multicolumn{1}{c}{-0.24}	                            & -0.60                           & \multicolumn{1}{c}{-1.80}	                & \multicolumn{1}{c}{-1.75}	                    & -1.68               & -1.01         & -1.31     \\
  \multicolumn{1}{l}{$\hspace{0.15cm} w/o$\ hl}             & \multicolumn{1}{c}{-2.25}                              & +0.90                           & \multicolumn{1}{c}{+0.42}                              & -0.75                           & \multicolumn{1}{c}{-0.74}	                & \multicolumn{1}{c}{-0.85}	                    & -0.79               & -0.20         & +0.56     \\  
  \multicolumn{1}{l}{$\hspace{0.15cm} w/o$\ CoT}            & \multicolumn{1}{c}{--        }                                     & --                                          & \multicolumn{1}{c}{--}                                             & --                                          & \multicolumn{1}{c}{-2.31}	                & \multicolumn{1}{c}{-2.35}	                    & -2.57               & -1.98         & -1.31      \\  \bottomrule
  \end{tabular}
  \caption{Ablation results on all datasets. ``$w/o$ TRC'', ``$w/o$ TCM'', and ``$w/o$ TCG'': HeGTa pre-trained without the specified objective. ``$w/o$ HG'': HeGTa pre-trained with homogeneous graphs. ``$w/o\ hl$'': HeGTa pre-trained with linking all neighboring cells directly instead of the heuristic linking strategy. 
  }
  \label{table: Ablation results}
\end{table*}

\footnotetext{For a fair comparison, all LLM-based methods use the open-source LLaMA2-7B~\cite{llama2} as the underlying model. More details of baselines are shown in Appendix.}

\noindent\textbf{General Setup across Evaluation Tasks.}
Following a typical $k$-shot setup~\cite{liu2021relative,wang2020graph} to validate the HeGTa's performance on few-shot TU, we randomly generate five 8-shot tasks for both training and validation on each dataset (i.e., 8 train/dev tables for CTC, 8 train/dev tables per class for TTC and 8 train/dev QA pairs for TQA). Each training task, paired with a corresponding validation task, is utilized to fine-tune the models optimally for subsequent testing. Table~\ref{table: Overall performance of the HeGTa.} presents the average performance on the test set across the five tasks.

\noindent\textbf{Evaluation Metrics.}
We adopt Macro-F1, Macro-F1 and accuracy as the evaluation metric for CTC, TTC and TQA, respectively.
Given the generative structure of LLMs, relying solely on exact matches might inaccurately categorize some correct responses (e.g., adding an answer with an additional period at the end). 
Therefore, we incorporate the semantic-match evaluator~\cite{Binding} for TQA, which preprocesses the outputs prior to evaluation.

\subsection{Results and Analysis}

Table~\ref{table: Overall performance of the HeGTa.} compares the performance of different methods on all datasets described in \S~\ref{sec: datasets}. 
From the results, We have the following observations:
1) HeGTa outperformed all baselines across 9 datasets, demonstrating its superiority over both non-LLM methods and LLMs with equivalent parameters. Additionally, as shown in Appendix, HeGTa demonstrates competitive performance in embedding-related tasks (i.e., CTC and TTC) even compared to SOTA GPT-4.
2) LLM-based symbolic methods excel in TQA but are inherently incapable of handling other table tasks due to their architectural design. Moreover, methods with 7B parameters struggle with reasoning and generating precise, executable code compared to those based on ChatGPT. For further details, see Appendix.
3) WTQ (Raw) represents the unprocessed version of WTQ (Flat), preserving the complexity of hierarchical table headers. The significant performance drop of other methods on WTQ (Raw) highlights HeGTa's ability to effectively handle relational tables with complex structures.
4) Compared to HeGTa, other LLM-based frameworks underperform due to the loss of table topological information, even with enhancements like TableLlama's extensive training on linearized tables or TableCoT's use of HTML, a format better suited for LLMs~\cite{tableformatsurvey}. This underscores the effectiveness of introducing GNNs in overcoming the limitations of linearized table representations, thereby unlocking the full potential of LLMs.

\subsubsection*{Strengths and Weaknesses of LLM-Based Models}

All QA samples of HiTab can be classified into lookup-based reasoning (L-R) and computation-based reasoning (C-R) (e.g., sum, max, count). Table~\ref{table: Analysis of HiTab} presents the performance of three LLM-based methods on two types of QA samples. 
TableCoT shows superior performance in L-R compared to C-R, aligning with the conclusion~\cite{chen2022program} that LLMs struggle with numerical calculations. 
COTable leverages SQL to mitigate the shortcomings of LLMs in numerical reasoning, leading to the highest accuracy in C-R. However, the hierarchical structure causes a decline in SQL performance. 
HeGTa addresses the issue of structural information loss of linearized tables. As a result, HeGTa achieves notable improvements, particularly in L-R samples. 
Additionally, without restricting to the few-shot setting, i.e., tuning the model with the entire training set, HeGTa consistently maintains top performance, as shown in Appendix.

\begin{table}[hbt]
  \centering
    \setlength{\tabcolsep}{1.1mm}
    \begin{tabular}{ccccccc}
    \toprule
    \multicolumn{1}{c}{\multirow{1}{*}{}}	                  &  \multicolumn{2}{c}{\multirow{1}{*}{\makecell{L-R}}}	                                                  & \multicolumn{2}{c}{\multirow{1}{*}{\makecell{C-R}}}               & \multicolumn{2}{c}{\multirow{1}{*}{\makecell{Total}}}                                                                  \\\cmidrule(lr){2-3} 	\cmidrule(lr){4-5} 	\cmidrule(lr){6-7}
    \multicolumn{1}{l}{}	                                  &  \# & \%	                                                                                                                    &  \# & \%	                                         	                                          &  \# & \%	                                         	                                                \\ 	
    
    \toprule
    \multicolumn{1}{l}{\multirow{1}{*}{\# Test set}}	      & 1133	 & 100.00        & 451           & 100.00    & 1584         & 100.00                                                                                      \\ \midrule[0.3pt]
    \multicolumn{1}{l}{\multirow{1}{*}{TableCoT}}	          & 612    & 54.02         & 195           & 43.24     & 806          & 47.55                                                                                         \\	\midrule[0.3pt]
    \multicolumn{1}{l}{\multirow{1}{*}{COTable}}            & 578 	 & 51.02         & 265           & 58.76     & 842          & 49.68                                                                                         \\  \midrule[0.3pt]
    \multicolumn{1}{l}{\multirow{1}{*}{HeGTa}}	            & 669    & 59.05         & 206           & 45.68     & 874          & 51.56                                                                                         \\	
    \bottomrule
    \end{tabular}
    \caption{Further results of the three methods on HiTab. ``\#'' represents the number of correctly answered samples, and ``\%'' denotes the proportion of correctly answered samples.} 
    \label{table: Analysis of HiTab}
  \end{table}

\subsection{Ablation Study}
We conduct ablation experiments to evaluate each component's effectiveness, as shown in Table~\ref{table: Ablation results}. 

We begin by examining the individual improvements contributed by each of the three self-training tasks. To evaluate their specific impact, we pre-train three models, each omitting one of the objectives: TRC, TCM, or TCG.
These three pre-training objectives improve the LLM's understanding of tabular structure. The TCM task most effectively aligns the tabular modality with the LLM's semantic space. The TRC task enhances performance on relational tables, while the TCG task boosts TQA performance but slightly reduces CTC performance.

We conducted ablation studies on CoT and our tabular graph construction method by removing the CoT text, converting tables into homogeneous rather than heterogeneous graphs, and replacing the heuristic linking (hl) strategy with a naive rule that links all neighboring cells. The results clearly demonstrate the significant impact of CoT on TQA tasks. The introduction of HGs benefits all tasks, with particularly notable improvements in TU with hierarchical structures. However, compared to the benefits of HGs, the hl strategy shows less pronounced gains for non-relational tables, possibly because the naive rule is more effective for aggregating and transmitting information in these tables.

\section{Conclusion}

In this paper, we introduce a novel framework, HeGTa, tailored for few-shot complex TU. The effectiveness of HeGTa is validated across multiple tabular datasets, accompanied by an in-depth ablation study to examine the impact of each component.
In future work, we plan to expand HeGTa's applicability to tables featuring more diverse layouts and to further improve HeGTa's performance in TQA tasks by integrating the program-aided paradigm into our approach.

\section{Acknowledgments}
This work is partially supported by National Nature Science Foundation of China under No. 62476058, by the project "Key Laboratory of rich-media Digital Publishing Content Organization and Knowledge Service Open Fund-Research on Knowledge-enhanced Training Techniques of Large Language Model" No. ZD2024-04/01. We thank the Big Data Computing Center of Southeast University for providing the facility support on the numerical calculations in this paper.

\nobibliography*
\bibliography{aaai25}

\begin{thebibliography}{44}
\providecommand{\natexlab}[1]{#1}

\bibitem[{Cao et~al.(2023)Cao, Chen, Liu, Wang, and Fried}]{api}
Cao, Y.; Chen, S.; Liu, R.; Wang, Z.; and Fried, D. 2023.
\newblock API-Assisted Code Generation for Question Answering on Varied Table Structures.
\newblock \emph{ArXiv}, abs/2310.14687.

\bibitem[{Chen(2022)}]{llmfew}
Chen, W. 2022.
\newblock Large Language Models are few(1)-shot Table Reasoners.
\newblock \emph{ArXiv}, abs/2210.06710.

\bibitem[{Chen et~al.(2022)Chen, Ma, Wang, and Cohen}]{chen2022program}
Chen, W.; Ma, X.; Wang, X.; and Cohen, W.~W. 2022.
\newblock Program of thoughts prompting: Disentangling computation from reasoning for numerical reasoning tasks.
\newblock \emph{arXiv preprint arXiv:2211.12588}.

\bibitem[{Chen et~al.(2020)Chen, Wang, Chen, Zhang, Wang, Li, Zhou, and Wang}]{tabfact}
Chen, W.; Wang, H.; Chen, J.; Zhang, Y.; Wang, H.; Li, S.; Zhou, X.; and Wang, W.~Y. 2020.
\newblock TabFact: A Large-scale Dataset for Table-based Fact Verification.
\newblock In \emph{International Conference on Learning Representations}.

\bibitem[{Cheng et~al.(2021{\natexlab{a}})Cheng, Dong, Cheng, Jia, Wu, Han, and Zhang}]{fortap}
Cheng, Z.; Dong, H.; Cheng, F.; Jia, R.; Wu, P.; Han, S.; and Zhang, D. 2021{\natexlab{a}}.
\newblock FORTAP: Using Formulas for Numerical-Reasoning-Aware Table Pretraining.
\newblock In \emph{Annual Meeting of the Association for Computational Linguistics}.

\bibitem[{Cheng et~al.(2021{\natexlab{b}})Cheng, Dong, Wang, Jia, Guo, Gao, Han, Lou, and Zhang}]{hitab}
Cheng, Z.; Dong, H.; Wang, Z.; Jia, R.; Guo, J.; Gao, Y.; Han, S.; Lou, J.-G.; and Zhang, D. 2021{\natexlab{b}}.
\newblock HiTab: A Hierarchical Table Dataset for Question Answering and Natural Language Generation.
\newblock \emph{ArXiv}, abs/2108.06712.

\bibitem[{Cheng et~al.(2022)Cheng, Xie, Shi, Li, Nadkarni, Hu, Xiong, Radev, Ostendorf, Zettlemoyer, Smith, and Yu}]{Binding}
Cheng, Z.; Xie, T.; Shi, P.; Li, C.; Nadkarni, R.; Hu, Y.; Xiong, C.; Radev, D.~R.; Ostendorf, M.; Zettlemoyer, L.; Smith, N.~A.; and Yu, T. 2022.
\newblock Binding Language Models in Symbolic Languages.
\newblock \emph{ArXiv}, abs/2210.02875.

\bibitem[{Chiang et~al.(2023)Chiang, Li, Lin, Sheng, Wu, Zhang, Zheng, Zhuang, Zhuang, Gonzalez et~al.}]{vicuna}
Chiang, W.-L.; Li, Z.; Lin, Z.; Sheng, Y.; Wu, Z.; Zhang, H.; Zheng, L.; Zhuang, S.; Zhuang, Y.; Gonzalez, J.~E.; et~al. 2023.
\newblock Vicuna: An open-source chatbot impressing gpt-4 with 90\%* chatgpt quality.
\newblock \emph{See https://vicuna. lmsys. org (accessed 14 April 2023)}.

\bibitem[{Deng et~al.(2020)Deng, Sun, Lees, Wu, and Yu}]{turl}
Deng, X.; Sun, H.; Lees, A.; Wu, Y.; and Yu, C. 2020.
\newblock TURL: Table Understanding through Representation Learning.
\newblock \emph{SIGMOD Rec.}, 51: 33--40.

\bibitem[{Devlin et~al.(2019)Devlin, Chang, Lee, and Toutanova}]{bert}
Devlin, J.; Chang, M.-W.; Lee, K.; and Toutanova, K. 2019.
\newblock BERT: Pre-training of Deep Bidirectional Transformers for Language Understanding.
\newblock \emph{ArXiv}, abs/1810.04805.

\bibitem[{Dong et~al.(2022)Dong, Cheng, He, Zhou, Zhou, Zhou, Liu, Han, and Zhang}]{dongsurvey}
Dong, H.; Cheng, Z.; He, X.; Zhou, M.; Zhou, A.; Zhou, F.; Liu, A.; Han, S.; and Zhang, D. 2022.
\newblock Table pre-training: A survey on model architectures, pre-training objectives, and downstream tasks.
\newblock \emph{arXiv preprint arXiv:2201.09745}.

\bibitem[{Du et~al.(2021)Du, Gao, Chen, Jia, Wang, Han, and Zhang}]{tabularnet}
Du, L.; Gao, F.; Chen, X.; Jia, R.; Wang, J.; Han, S.; and Zhang, D. 2021.
\newblock TabularNet: A Neural Network Architecture for Understanding Semantic Structures of Tabular Data.
\newblock \emph{Proceedings of the 27th ACM SIGKDD Conference on Knowledge Discovery \& Data Mining}.

\bibitem[{Ghasemi-Gol and Szekely(2018)}]{wcc}
Ghasemi-Gol, M.; and Szekely, P.~A. 2018.
\newblock TabVec: Table Vectors for Classification of Web Tables.
\newblock \emph{ArXiv}, abs/1802.06290.

\bibitem[{Herzig et~al.(2020)Herzig, Nowak, M{\"u}ller, Piccinno, and Eisenschlos}]{tapas}
Herzig, J.; Nowak, P.~K.; M{\"u}ller, T.; Piccinno, F.; and Eisenschlos, J.~M. 2020.
\newblock TaPas: Weakly Supervised Table Parsing via Pre-training.
\newblock \emph{ArXiv}, abs/2004.02349.

\bibitem[{Hu et~al.(2021)Hu, Shen, Wallis, Allen-Zhu, Li, Wang, Wang, and Chen}]{lora}
Hu, E.~J.; Shen, Y.; Wallis, P.; Allen-Zhu, Z.; Li, Y.; Wang, S.; Wang, L.; and Chen, W. 2021.
\newblock Lora: Low-rank adaptation of large language models.
\newblock \emph{arXiv preprint arXiv:2106.09685}.

\bibitem[{Hu et~al.(2020)Hu, Dong, Wang, and Sun}]{HGT}
Hu, Z.; Dong, Y.; Wang, K.; and Sun, Y. 2020.
\newblock Heterogeneous graph transformer.
\newblock In \emph{Proceedings of the web conference 2020}, 2704--2710.

\bibitem[{Jia et~al.(2023)Jia, Guo, Jin, Yan, Du, Ma, Stankovic, Lozajic, Zoranovic, Ilic, Han, and Zhang}]{getpt}
Jia, R.; Guo, H.; Jin, X.; Yan, C.; Du, L.; Ma, X.; Stankovic, T.; Lozajic, M.; Zoranovic, G.; Ilic, I.; Han, S.; and Zhang, D. 2023.
\newblock GetPt: Graph-enhanced General Table Pre-training with Alternate Attention Network.
\newblock \emph{Proceedings of the 29th ACM SIGKDD Conference on Knowledge Discovery and Data Mining}.

\bibitem[{Jin et~al.(2023{\natexlab{a}})Jin, Wang, Ma, Chu, Zhang, Shi, Chen, Liang, Li, Pan, and Wen}]{timellm}
Jin, M.; Wang, S.; Ma, L.; Chu, Z.; Zhang, J.~Y.; Shi, X.~L.; Chen, P.-Y.; Liang, Y.; Li, Y.-F.; Pan, S.; and Wen, Q. 2023{\natexlab{a}}.
\newblock Time-LLM: Time Series Forecasting by Reprogramming Large Language Models.
\newblock \emph{ArXiv}, abs/2310.01728.

\bibitem[{Jin et~al.(2023{\natexlab{b}})Jin, Wang, Tan, Chen, Qi, and Hao}]{tabprompt}
Jin, R.; Wang, J.; Tan, W.; Chen, Y.; Qi, G.; and Hao, W. 2023{\natexlab{b}}.
\newblock TabPrompt: Graph-based Pre-training and Prompting for Few-shot Table Understanding.
\newblock In \emph{Conference on Empirical Methods in Natural Language Processing}.

\bibitem[{Li and Liang(2021)}]{softprompt}
Li, X.~L.; and Liang, P. 2021.
\newblock Prefix-Tuning: Optimizing Continuous Prompts for Generation.
\newblock \emph{Proceedings of the 59th Annual Meeting of the Association for Computational Linguistics and the 11th International Joint Conference on Natural Language Processing (Volume 1: Long Papers)}, abs/2101.00190.

\bibitem[{Liu et~al.(2023)Liu, Li, Wu, and Lee}]{llava}
Liu, H.; Li, C.; Wu, Q.; and Lee, Y.~J. 2023.
\newblock Visual Instruction Tuning.
\newblock \emph{ArXiv}, abs/2304.08485.

\bibitem[{Liu et~al.(2021)Liu, Fang, Liu, and Hoi}]{liu2021relative}
Liu, Z.; Fang, Y.; Liu, C.; and Hoi, S.~C. 2021.
\newblock Relative and absolute location embedding for few-shot node classification on graph.
\newblock In \emph{Proceedings of the AAAI conference on artificial intelligence}, 4267--4275.

\bibitem[{Lu et~al.(2024)Lu, Zhang, Zhang, and Chen}]{lu2024large}
Lu, W.; Zhang, J.; Zhang, J.; and Chen, Y. 2024.
\newblock Large language model for table processing: A survey.
\newblock \emph{arXiv preprint arXiv:2402.05121}.

\bibitem[{Ma, Liu, and Zuo(2023)}]{sshg1}
Ma, S.; Liu, J.-w.; and Zuo, X. 2023.
\newblock Self-supervised learning for heterogeneous graph via structure information based on metapath.
\newblock \emph{Applied Soft Computing}, 143: 110388.

\bibitem[{Ouyang et~al.(2022)Ouyang, Wu, Jiang, Almeida, Wainwright, Mishkin, Zhang, Agarwal, Slama, Ray, Schulman, Hilton, Kelton, Miller, Simens, Askell, Welinder, Christiano, Leike, and Lowe}]{instruct2}
Ouyang, L.; Wu, J.; Jiang, X.; Almeida, D.; Wainwright, C.~L.; Mishkin, P.; Zhang, C.; Agarwal, S.; Slama, K.; Ray, A.; Schulman, J.; Hilton, J.; Kelton, F.; Miller, L.~E.; Simens, M.; Askell, A.; Welinder, P.; Christiano, P.~F.; Leike, J.; and Lowe, R.~J. 2022.
\newblock Training language models to follow instructions with human feedback.
\newblock \emph{ArXiv}, abs/2203.02155.

\bibitem[{Pasupat and Liang(2015)}]{wtq}
Pasupat, P.; and Liang, P. 2015.
\newblock Compositional Semantic Parsing on Semi-Structured Tables.
\newblock In \emph{Annual Meeting of the Association for Computational Linguistics}.

\bibitem[{Reimers and Gurevych(2019)}]{sentencebert}
Reimers, N.; and Gurevych, I. 2019.
\newblock Sentence-BERT: Sentence Embeddings using Siamese BERT-Networks.
\newblock In \emph{Proceedings of the 2019 Conference on Empirical Methods in Natural Language Processing}. Association for Computational Linguistics.

\bibitem[{Sui et~al.(2024)Sui, Zhou, Zhou, Han, and Zhang}]{tableformatsurvey}
Sui, Y.; Zhou, M.; Zhou, M.; Han, S.; and Zhang, D. 2024.
\newblock Table meets llm: Can large language models understand structured table data? a benchmark and empirical study.
\newblock In \emph{Proceedings of the 17th ACM International Conference on Web Search and Data Mining}, 645--654.

\bibitem[{Tang et~al.(2023)Tang, Yang, Wei, Shi, Su, Cheng, Yin, and Huang}]{graphgpt}
Tang, J.; Yang, Y.; Wei, W.; Shi, L.; Su, L.; Cheng, S.; Yin, D.; and Huang, C. 2023.
\newblock GraphGPT: Graph Instruction Tuning for Large Language Models.
\newblock \emph{ArXiv}, abs/2310.13023.

\bibitem[{Touvron et~al.(2023)Touvron, Martin, Stone, Albert, Almahairi, Babaei, Bashlykov, Batra, Bhargava, Bhosale, Bikel, Blecher, Ferrer, Chen, Cucurull, Esiobu, Fernandes, Fu, Fu, Fuller, Gao, Goswami, Goyal, Hartshorn, Hosseini, Hou, Inan, Kardas, Kerkez, Khabsa, Kloumann, Korenev, Koura, Lachaux, Lavril, Lee, Liskovich, Lu, Mao, Martinet, Mihaylov, Mishra, Molybog, Nie, Poulton, Reizenstein, Rungta, Saladi, Schelten, Silva, Smith, Subramanian, Tan, Tang, Taylor, Williams, Kuan, Xu, Yan, Zarov, Zhang, Fan, Kambadur, Narang, Rodriguez, Stojnic, Edunov, and Scialom}]{llama2}
Touvron, H.; Martin, L.; Stone, K.~R.; Albert, P.; Almahairi, A.; Babaei, Y.; Bashlykov, N.; Batra, S.; Bhargava, P.; Bhosale, S.; Bikel, D.~M.; Blecher, L.; Ferrer, C.~C.; Chen, M.; Cucurull, G.; Esiobu, D.; Fernandes, J.; Fu, J.; Fu, W.; Fuller, B.; Gao, C.; Goswami, V.; Goyal, N.; Hartshorn, A.~S.; Hosseini, S.; Hou, R.; Inan, H.; Kardas, M.; Kerkez, V.; Khabsa, M.; Kloumann, I.~M.; Korenev, A.~V.; Koura, P.~S.; Lachaux, M.-A.; Lavril, T.; Lee, J.; Liskovich, D.; Lu, Y.; Mao, Y.; Martinet, X.; Mihaylov, T.; Mishra, P.; Molybog, I.; Nie, Y.; Poulton, A.; Reizenstein, J.; Rungta, R.; Saladi, K.; Schelten, A.; Silva, R.; Smith, E.~M.; Subramanian, R.; Tan, X.; Tang, B.; Taylor, R.; Williams, A.; Kuan, J.~X.; Xu, P.; Yan, Z.; Zarov, I.; Zhang, Y.; Fan, A.; Kambadur, M.; Narang, S.; Rodriguez, A.; Stojnic, R.; Edunov, S.; and Scialom, T. 2023.
\newblock Llama 2: Open Foundation and Fine-Tuned Chat Models.
\newblock \emph{ArXiv}, abs/2307.09288.

\bibitem[{Wang et~al.(2021)Wang, Sun, Chen, Pujara, and Szekely}]{GTR}
Wang, F.; Sun, K.; Chen, M.; Pujara, J.; and Szekely, P. 2021.
\newblock Retrieving complex tables with multi-granular graph representation learning.
\newblock In \emph{Proceedings of the 44th International ACM SIGIR Conference on Research and Development in Information Retrieval}, 1472--1482.

\bibitem[{Wang et~al.(2020{\natexlab{a}})Wang, Shen, Yang, Quan, and Wang}]{rgat}
Wang, K.; Shen, W.; Yang, Y.; Quan, X.; and Wang, R. 2020{\natexlab{a}}.
\newblock Relational Graph Attention Network for Aspect-based Sentiment Analysis.
\newblock In \emph{ACL}.

\bibitem[{Wang et~al.(2020{\natexlab{b}})Wang, Luo, Ding, Zhang, Li, and Zheng}]{wang2020graph}
Wang, N.; Luo, M.; Ding, K.; Zhang, L.; Li, J.; and Zheng, Q. 2020{\natexlab{b}}.
\newblock Graph few-shot learning with attribute matching.
\newblock In \emph{Proceedings of the 29th ACM International Conference on Information \& Knowledge Management}, 1545--1554.

\bibitem[{Wang et~al.(2022)Wang, Kordi, Mishra, Liu, Smith, Khashabi, and Hajishirzi}]{instruct1}
Wang, Y.; Kordi, Y.; Mishra, S.; Liu, A.; Smith, N.~A.; Khashabi, D.; and Hajishirzi, H. 2022.
\newblock Self-Instruct: Aligning Language Models with Self-Generated Instructions.
\newblock In \emph{Annual Meeting of the Association for Computational Linguistics}.

\bibitem[{Wang et~al.(2020{\natexlab{c}})Wang, Dong, Jia, Li, Fu, Han, and Zhang}]{tuta}
Wang, Z.; Dong, H.; Jia, R.; Li, J.; Fu, Z.; Han, S.; and Zhang, D. 2020{\natexlab{c}}.
\newblock TUTA: Tree-based Transformers for Generally Structured Table Pre-training.
\newblock \emph{Proceedings of the 27th ACM SIGKDD Conference on Knowledge Discovery \& Data Mining}.

\bibitem[{Wang et~al.(2024)Wang, Zhang, Li, Eisenschlos, Perot, Wang, Miculicich, Fujii, Shang, Lee et~al.}]{chain-of-table}
Wang, Z.; Zhang, H.; Li, C.-L.; Eisenschlos, J.~M.; Perot, V.; Wang, Z.; Miculicich, L.; Fujii, Y.; Shang, J.; Lee, C.-Y.; et~al. 2024.
\newblock Chain-of-table: Evolving tables in the reasoning chain for table understanding.
\newblock \emph{arXiv preprint arXiv:2401.04398}.

\bibitem[{Wei et~al.(2022)Wei, Wang, Schuurmans, Bosma, Xia, Chi, Le, Zhou et~al.}]{wei2022chain}
Wei, J.; Wang, X.; Schuurmans, D.; Bosma, M.; Xia, F.; Chi, E.; Le, Q.~V.; Zhou, D.; et~al. 2022.
\newblock Chain-of-thought prompting elicits reasoning in large language models.
\newblock \emph{Advances in neural information processing systems}, 35: 24824--24837.

\bibitem[{Yang et~al.(2022)Yang, Guan, Wang, Zhao, Xu, Lu, and Huang}]{sshg2}
Yang, Y.; Guan, Z.; Wang, Z.; Zhao, W.; Xu, C.; Lu, W.; and Huang, J. 2022.
\newblock Self-supervised heterogeneous graph pre-training based on structural clustering.
\newblock \emph{Advances in Neural Information Processing Systems}, 35: 16962--16974.

\bibitem[{Ye et~al.(2023)Ye, Hui, Yang, Li, Huang, and Li}]{dater}
Ye, Y.; Hui, B.; Yang, M.; Li, B.; Huang, F.; and Li, Y. 2023.
\newblock Large language models are versatile decomposers: Decomposing evidence and questions for table-based reasoning.
\newblock In \emph{Proceedings of the 46th International ACM SIGIR Conference on Research and Development in Information Retrieval}, 174--184.

\bibitem[{Yin et~al.(2020)Yin, Neubig, tau Yih, and Riedel}]{tabert}
Yin, P.; Neubig, G.; tau Yih, W.; and Riedel, S. 2020.
\newblock TaBERT: Pretraining for Joint Understanding of Textual and Tabular Data.
\newblock \emph{ArXiv}, abs/2005.08314.

\bibitem[{Zhang et~al.(2023{\natexlab{a}})Zhang, Yue, Li, and Sun}]{tablellama}
Zhang, T.; Yue, X.; Li, Y.; and Sun, H. 2023{\natexlab{a}}.
\newblock TableLlama: Towards Open Large Generalist Models for Tables.
\newblock \emph{ArXiv}, abs/2311.09206.

\bibitem[{Zhang et~al.(2023{\natexlab{b}})Zhang, Henkel, Floratou, Cahoon, Deep, and Patel}]{ReAcTable}
Zhang, Y.; Henkel, J.; Floratou, A.; Cahoon, J.; Deep, S.; and Patel, J.~M. 2023{\natexlab{b}}.
\newblock ReAcTable: Enhancing ReAct for Table Question Answering.
\newblock \emph{ArXiv}, abs/2310.00815.

\bibitem[{Zhang et~al.(2022)Zhang, Zhang, Li, and Smola}]{zhang2022automatic}
Zhang, Z.; Zhang, A.; Li, M.; and Smola, A. 2022.
\newblock Automatic chain of thought prompting in large language models.
\newblock \emph{arXiv preprint arXiv:2210.03493}.

\bibitem[{Zheng et~al.(2023)Zheng, Hao, Jiang, Lin, Lyu, She, and Wang}]{imtqa}
Zheng, M.; Hao, Y.; Jiang, W.-J.; Lin, Z.; Lyu, Y.; She, Q.; and Wang, W. 2023.
\newblock IM-TQA: A Chinese Table Question Answering Dataset with Implicit and Multi-type Table Structures.
\newblock In \emph{Annual Meeting of the Association for Computational Linguistics}.

\end{thebibliography}

\end{document}